\documentclass[lettersize,twoside,journal]{IEEEtran}
\usepackage{graphicx}
\usepackage{amssymb}
\usepackage{amsmath}
\usepackage{cases}
\usepackage{cite}
\usepackage{color}
\usepackage{stfloats}
\usepackage{subfigure}
\usepackage{booktabs}
\usepackage{balance}
\usepackage{braket}
\usepackage{acro}
\usepackage{multirow}
\usepackage{makecell}

\usepackage{xcolor}
\usepackage{braket}
\usepackage{bbm}
\usepackage{pifont} 

\DeclareMathAlphabet{\mathpzc}{OT1}{pzc}{m}{it}

\DeclareMathAlphabet{\mathitsf}{OML}{cmbr}{m}{it}
\DeclareMathAlphabet{\mathsf}{OT1}{cmbr}{m}{n}

\definecolor{darkgreen}{RGB}{0,128,0}

\DeclareAcronym{1D}{
short = 1D,
long = one-dimensional
}

\DeclareAcronym{2D}{
short = 2D,
long = two-dimensional
}

\DeclareAcronym{3D}{
short = 3D,
long = three-dimensional
}

\DeclareAcronym{5G}{
short = 5G,
long = fifth-generation
}

\DeclareAcronym{6G}{
short = 6G,
long = sixth-generation
}

\DeclareAcronym{AI}{
short = AI,
long = artificial intelligence
}

\DeclareAcronym{ANN}{
short = ANN,
long = artificial neural network
}

\DeclareAcronym{AR}{
short = AR,
long = augmented reality
}

\DeclareAcronym{AO}{
short = AO,
long = absorptive object
}

\DeclareAcronym{AWGN}{
short = AWGN,
long = additive white Gaussian noise
}

\DeclareAcronym{BEP}{
short = BEP,
long = bit error probability
}

\DeclareAcronym{BLIP}{
short = BLIP,
long = bootstrapping language-image pretraining
}

\DeclareAcronym{BPSK}{
short = BPSK,
long = binary phase-shift keying
}

\DeclareAcronym{BS}{
short = BS,
long = base station
}

\DeclareAcronym{BSM}{
short = BSM,
long = Bell state measurement
}

\DeclareAcronym{CKA}{
short = CKA,
long = conference key agreement
}

\DeclareAcronym{CNN}{
short = CNN,
long = convolutional neural network
}

\DeclareAcronym{CNOT}{
short = CNOT,
long = controlled-NOT
}

\DeclareAcronym{CPS}{
short = CPS,
long = cyber-physical system
}

\DeclareAcronym{CPTP}{
short = CPTP,
long = completely positive trace-preserving
}

\DeclareAcronym{CPU}{
short = CPU,
long = central processing unit
}

\DeclareAcronym{CSI}{
short = CSI,
long = channel state information
}

\DeclareAcronym{DNN}{
short = DNN,
long = deep neural network
}

\DeclareAcronym{DL}{
short = DL,
long = deep learning
}

\DeclareAcronym{DQL}{
short = DQL,
long = deep Q-learning
}

\DeclareAcronym{DRL}{
short = DRL,
alt = deep RL,
long = deep reinforcement learning
}

\DeclareAcronym{DT}{
short = DT,
long = digital twin
}

\DeclareAcronym{GAN}{
short = GAN,
long = generative adversarial network
}

\DeclareAcronym{GEO}{
short = GEO,
long = geostationary Earth orbit
}

\DeclareAcronym{GHZ}{
short = GHZ,
long = Greenberger--Horne--Zeilinger
}

\DeclareAcronym{GPS}{
short = GPS,
long = global positioning system
}

\DeclareAcronym{GPT}{
short = GPT,
long = generative pretrained transformer
}

\DeclareAcronym{GPU}{
short = GPU,
long = graphics processing unit 
}

\DeclareAcronym{GUS}{
short = GUS,
long = geometrically uniform symmetry
}

\DeclareAcronym{HQC}{
short = HQC,
long = hybrid quantum-classical
}

\DeclareAcronym{IoT}{
short = IoT,
long = Internet of Things
}

\DeclareAcronym{IIoT}{
short = IIoT,
alt = industrial IoT,
long = industrial Internet of Things
}

\DeclareAcronym{ISAC}{
short = ISAC,
long = integrated sensing and communication
}

\DeclareAcronym{KPI}{
short = KPI,
long = key performance indicator
}

\DeclareAcronym{LEO}{
short = LEO,
long = low Earth orbit
}

\DeclareAcronym{LLaMA}{
short = LLaMA,
alt = large language model Meta AI,
long = LLM Meta AI
}

\DeclareAcronym{LLM}{
short = LLM,
long = large language model
}

\DeclareAcronym{LPIPS}{
short = LPIPS,
long = learned perceptual image patch similarity
}

\DeclareAcronym{LOCC}{
short = LOCC,
long = local operations and classical communication
}

\DeclareAcronym{LSTM}{
short = LSTM,
long = long short-term memory
}

\DeclareAcronym{MEO}{
short = MEO,
long = medium Earth orbit
}

\DeclareAcronym{MIMO}{
short = MIMO,
long = multiple-input multiple-output
}

\DeclareAcronym{mmWave}{
short = mmWave,
long = millimeter-wave
}

\DeclareAcronym{ML}{
short = ML,
long = machine learning
}

\DeclareAcronym{MLP}{
short = MLP,
long = multi-layer perceptron
}

\DeclareAcronym{MPA}{
short = MPA,
long = mean pixel accuracy
}

\DeclareAcronym{MR}{
short = MR,
long = mixed reality
}

\DeclareAcronym{MSE}{
short = MSE,
long = mean squared error
}

\DeclareAcronym{NISQ}{
short = NISQ,
long = noisy intermediate-scale quantum
}

\DeclareAcronym{NMSE}{
short = NMSE,
alt = normalized MSE,
long = normalized mean squared error
}

\DeclareAcronym{NLP}{
short = NLP,
long = natural language processing
}

\DeclareAcronym{NTN}{
short = NTN,
long = non-terrestrial network
}

\DeclareAcronym{PISQ}{
short = PISQ,
long = perfect intermediate-scale quantum
}

\DeclareAcronym{POVM}{
short = POVM,
long = positive operator-valued measure
}

\DeclareAcronym{PQC}{
short = PQC,
long = parametrized quantum circuit
}

\DeclareAcronym{QAA}{
short = QAA,
long = quantum anonymous authentication
}

\DeclareAcronym{QAB}{
short = QAB,
long = quantum anonymous broadcast
}

\DeclareAcronym{QAC}{
short = QAC,
long = quantum anonymous communication
}

\DeclareAcronym{QACKA}{
short = QA-CKA,
alt = quantum anonymous CKA,
long = quantum anonymous conference key agreement
}

\DeclareAcronym{QACD}{
short = QACD,
long = quantum anonymous collision detection
}

\DeclareAcronym{QAE}{
short = QAE,
long = quantum anonymous entanglement
}

\DeclareAcronym{QAIA}{
short = QAIA,
long = quantum anonymous identity authentication
}

\DeclareAcronym{QAIR}{
short = QAIR,
long = quantum anonymous information retrieval
}

\DeclareAcronym{QAM}{
short = QAM,
long = quadrature amplitude modulation
}

\DeclareAcronym{QAN}{
short = QAN,
long = quantum anonymous network
}

\DeclareAcronym{qAN}{
short = QAN,
long = quantum anonymous notification
}

\DeclareAcronym{QANO}{
short = QANO,
long = quantum anonymous notification
}

\DeclareAcronym{QAP}{
short = QAP,
long = quantum anonymous publication
}

\DeclareAcronym{QAR}{
short = QAR,
long = quantum anonymous ranking
}

\DeclareAcronym{QAS}{
short = QAS,
long = quantum anonymous sensing
}

\DeclareAcronym{QAT}{
short = QAT,
long = quantum anonymous teleportation
}

\DeclareAcronym{QAV}{
short = QAV,
long = quantum anonymous voting
}

\DeclareAcronym{QCRB}{
short = QCRB,
long = quantum Cram\'er--Rao bound
}

\DeclareAcronym{QEC}{
short = QEC,
long = quantum error correction
}

\DeclareAcronym{QFI}{
short = QFI,
long = quantum Fisher information
}

\DeclareAcronym{QFIM}{
short = QFIM,
alt = QFI matrix,
long = quantum Fisher information matrix
}

\DeclareAcronym{QFT}{
short = QFT,
long = quantum Fourier transform
}

\DeclareAcronym{QIoT}{
short = QIoT,
alt = quantum IoT,
long = quantum Internet of Things
}

\DeclareAcronym{QKD}{
short = QKD,
long = quantum key distribution
}

\DeclareAcronym{QML}{
short = QML,
alt = quantum ML,
long = quantum machine learning
}

\DeclareAcronym{QNN}{
short = QNN,
long = quantum neural network
}

\DeclareAcronym{QOC}{
short = QOC,
long = quantum optimal control
}

\DeclareAcronym{QPU}{
short = QPU,
long = quantum processing unit
}

\DeclareAcronym{QSC}{
short = QSC,
alt = quantum SC,
long = quantum semantic communication
}

\DeclareAcronym{QSN}{
short = QSN,
long = quantum sensing network
}

\DeclareAcronym{ReLU}{
short = ReLU,
long = rectified linear unit
}

\DeclareAcronym{ResNet}{
short = ResNet,
long = residual network
}

\DeclareAcronym{RGB}{
short = RGB,
long = red-green-blue
}

\DeclareAcronym{RL}{
short = RL,
long = reinforcement learning
}

\DeclareAcronym{RNN}{
short = RNN,
long = recurrent neural network
}

\DeclareAcronym{RSA}{
short = RSA,
long = Rivest--Shamir--Adleman
}

\DeclareAcronym{SAM}{
short = SAM,
long = segment anything model
}

\DeclareAcronym{SC}{
short = SC,
long = semantic communication
}

\DeclareAcronym{SEP}{
short = SEP,
long = symbol error probability
}

\DeclareAcronym{SNR}{
short = SNR,
long = signal-to-noise ratio
}

\DeclareAcronym{SRM}{
short = SRM,
long = square-root measurement
}

\DeclareAcronym{SQL}{
short = SQL,
long = standard quantum limit
}

\DeclareAcronym{SVM}{
short = SVM,
long = support vector machine
}

\DeclareAcronym{SwinT}{
short = SwinT,
long = swin (shifted window) transformer
}

\DeclareAcronym{THz}{
short = THz,
long = terahertz
}

\DeclareAcronym{TN}{
short = TN,
long = terrestrial network
}

\DeclareAcronym{UAV}{
short = UAV,
long = unmanned aerial vehicle
}

\DeclareAcronym{VAE}{
short = VAE,
long = variational autoencoder
}

\DeclareAcronym{V2I}{
short = V2I,
long = vehicle-to-infrastructure
}

\DeclareAcronym{V2X}{
short = V2X,
long = vehicle-to-everything
}

\DeclareAcronym{VFM}{
short = VFM,
long = vision foundation model
}

\DeclareAcronym{ViT}{
short = ViT,
long = vision transformer
}

\DeclareAcronym{VQA}{
short = VQA,
long = variational quantum algorithm
}

\DeclareAcronym{VQC}{
short = VQC,
long = variational quantum circuit
}

\DeclareAcronym{VQS}{
short = VQS,
long = variational quantum sensing
}

\DeclareAcronym{VR}{
short = VR,
long = virtual reality
}

\DeclareAcronym{YOLO}{
short = YOLO,
long = you only look once
}

\DeclareAcronym{AIGC}{
  short = AIGC,
  long = AI-generated content
}

\DeclareAcronym{CLIP}{
  short = CLIP,
  long = contrastive language-image pretraining
}

\DeclareAcronym{CLIP-S}{
  short = CLIP-S,
  long = contrastive language-image pretraining score
}

\DeclareAcronym{IoU}{
  short = IoU,
  long = intersection over union
}

\DeclareAcronym{JSCC}{
  short = JSCC,
  long = joint source-channel coding
}

\DeclareAcronym{KB}{
  short = KB,
  long = knowledge base
}

\DeclareAcronym{LDM}{
  short = LDM,
  long = latent diffusion model
}

\DeclareAcronym{OFDM}{
  short = OFDM,
  long = orthogonal frequency-division multiplexing
}

\DeclareAcronym{PSNR}{
  short = PSNR,
  long = peak signal-to-noise ratio
}

\DeclareAcronym{SDR}{
  short = SDR,
  long = software-defined radio
}

\DeclareAcronym{VGG}{
  short = VGG,
  long = visual geometry group
}

\begin{document}
\bstctlcite{IEEEexample:BSTcontrol}

\title{
GenSC-6G: A Prototype Testbed for Integrated Generative AI, Quantum, and Semantic Communication
} 

\author{
Brian~E.~Arfeto,
Shehbaz~Tariq,
Uman~Khalid,
Trung~Q.~Duong,~\IEEEmembership{Fellow,~IEEE},
and
Hyundong~Shin,~\IEEEmembership{Fellow,~IEEE}
\thanks{
B.~E.~Arfeto,
S.~Tariq,
U.~Khalid,
and 
H.~Shin (corresponding author)
are with
Kyung Hee University,
Korea; 
T.~Q.~Duong is with 
Memorial University, Canada,
and Queen’s University Belfast, UK.
}
}


\markboth{
Submitted for Publication in IEEE Communications Magazine}{ 
Arfeto \textit{\MakeLowercase{et al.}}:
GenSC-6G: A Prototype Testbed for Integrated Generative AI, Quantum, and Semantic Communication
}

\maketitle

\begin{abstract}

We introduce a prototyping testbed---GenSC-6G---developed to generate a comprehensive dataset that supports the integration of generative \ac{AI}, quantum computing, and semantic communication for emerging \ac{6G} applications. The GenSC-6G dataset is designed with noise-augmented synthetic data optimized for semantic decoding, classification, and localization tasks, significantly enhancing flexibility for diverse AI-driven communication applications. This adaptable prototype supports seamless modifications across baseline models, communication modules, and goal-oriented decoders. Case studies demonstrate its application in lightweight classification, semantic upsampling, and edge-based language inference under noise conditions. The GenSC-6G dataset serves as a scalable and robust resource for developing goal-oriented communication systems tailored to the growing demands of \ac{6G} networks.

\end{abstract}

\begin{IEEEkeywords}

6G testbed,
generative AI,
hybrid quantum-classical computing,
semantic communication.
 
\end{IEEEkeywords}

\bstctlcite{refs-control}
\section{Introduction}
\label{sec:1}

\acresetall		

\begin{figure}[t]
\centering
\includegraphics[width=0.93\linewidth]{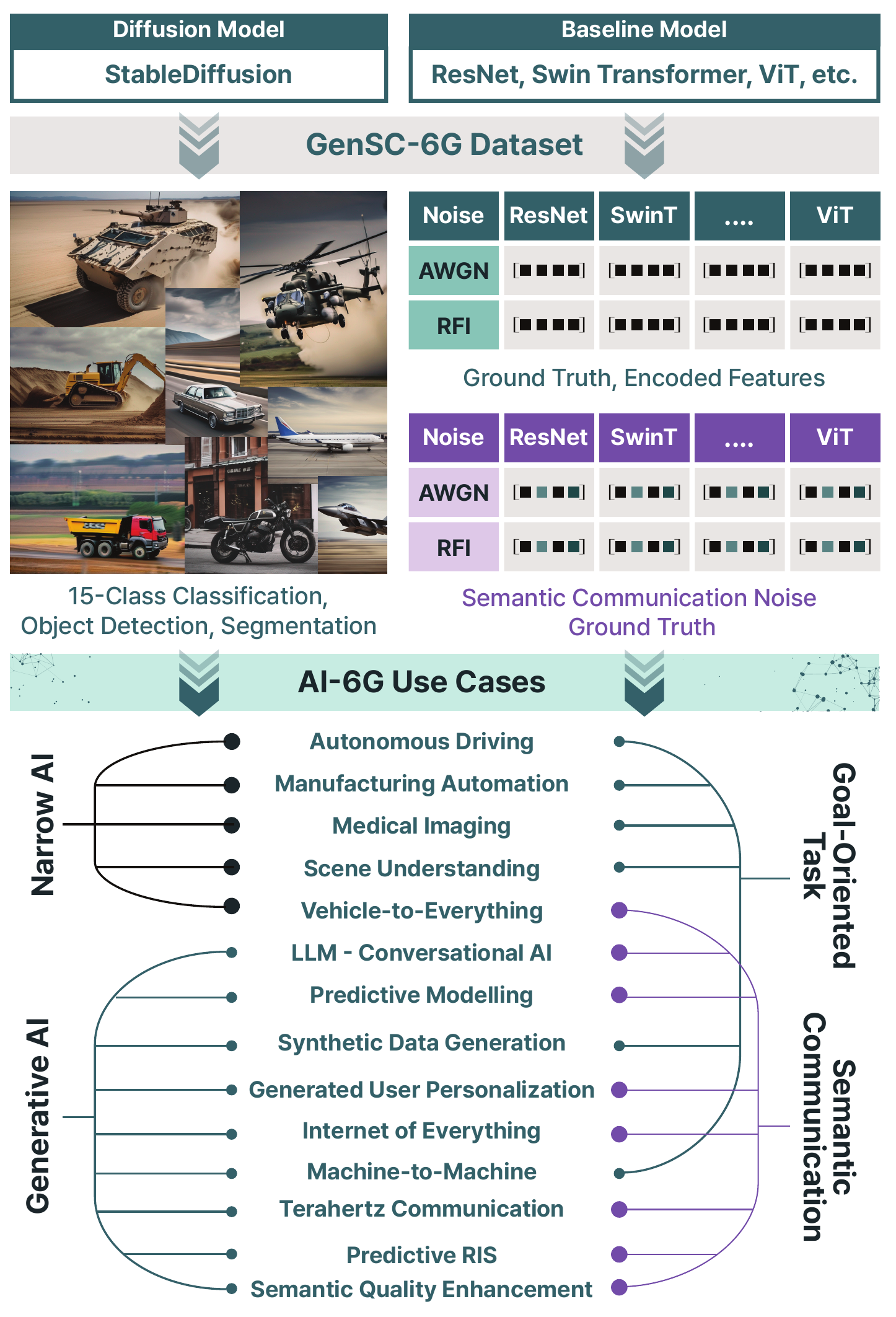}
\caption{
GenSC-6G dataset structuress. The dataset consists of the ground-truth data, encoded features, and additive noise. The AI-6G use cases span narrow AI and generative AI applications over SC and goal-oriented tasks in multiple fields. ResNet, ViT, SwinT, AWGN, RFI, LLM, and RIS stand for residual network, vision transformer, swin transformer, additive white Gaussian noise, radio frequency interference, large language model, and reconfigurable intelligent surface, respectively.
}
\vspace{-10pt}
\label{fig:1}
\end{figure}

\IEEEPARstart{T}{he} integration of generative \ac{AI} with \ac{SC} marks a transformative paradigm shift in communications, transitioning from basic data transmission to goal-oriented, context-aware information exchange \cite{CSDHP:24:IEEE_O_CSTO, LLCCHZZ:24:IEEE_O_CSTO}. By leveraging advanced \ac{AI} and foundation models \cite{FLYWSGX:2022:NC, XZC:2023:IEEE_J_PAMI}, these systems enhance efficiency and adaptability, tailoring transmissions to align with specific communicative goals. At its core, the \ac{SC} employs a \ac{KB} with semantic encoders and decoders, prioritizing context and intent over raw data \cite{GQHYC:2023:IEEE_J_JSAC}. This innovative approach enables ultra-efficient compression, making it ideal for applications such as \ac{IoT}, cloud services, autonomous systems, and other cutting-edge \ac{6G} use cases, as illustrated in Fig.~\ref{fig:1}.

Recent advancements have demonstrated the successful integration of advanced \ac{AI} and \ac{SC} systems, paving the way for more adaptive and intelligent communication networks \cite{JDPWYPY:24:IEEE_M_COM}. Building on this progress, numerous studies have explored how generative \ac{AI} can enhance data generation and transmission quality in \ac{6G} networks \cite{JPDWYPND:24:IEEE_M_WC, TKA:24:IEEE_M_WC}. For example, generative \ac{AI} has been utilized at the network edge to improve visual data transmission quality by leveraging multimodal data inputs \cite{ZYHCH:24:IEEE_J_CCN}. Similarly, integrating \ac{AI} into communication devices has facilitated support for diverse data formats and translation tasks \cite{LDSNKZI:24:IEEE_J_CCN}. Moreover, advanced \ac{AI} techniques have been embedded in decoders to allow the generation of new information on the receiver end, further enhancing the potential of \ac{AIGC} \cite{YXMQZDT:24:arXiv}. Generative models play a crucial role in creating and updating the \ac{KB} within \ac{SC} systems by dynamically generating refined knowledge representations \cite{RZXCZSZC:24:IEEE_M_WC}. This \ac{KB} serves as a repository for learned semantics, enabling efficient encoding and decoding while minimizing raw data transmission. Constructing a robust \ac{KB}, however, requires extensive labeled datasets for effective model training. 

While testbeds, such as DeepSense 6G \cite{ACOHMDS:23:IEEE_M_COM}, offer realistic environments for sub-\ac{THz} beam prediction, their high cost and limited adaptability pose scalability challenges. Generative \ac{AI} addresses these limitations by producing realistic synthetic datasets, reducing the manual data collection burden, and enhancing scalability across domains.
In addition to scalability, sustainability is critical in communication systems, especially for \ac{6G} applications on edge devices. \Ac{HQC} processing and model pruning contribute to this goal by offloading computationally intensive tasks and enabling faster inference \cite{TAK:24:IEEE_J_IOT}. Furthermore, recent work in quantum \ac{SC} has highlighted robustness and security advantages of quantum integration, further broadening the potential of sustainable \ac{SC} frameworks \cite{KUFDDS:23:IEEE_M_WC}. To address these challenges, this paper offers the following contributions:

\begin{itemize}
   
\item 
\textbf{Adaptable \ac{SC} Framework}: A flexible prototype that supports modifications to baseline models, communication modules, and decoders, enabling customization for diverse communication needs.    

\item 
\textbf{Generative \ac{AI}-Driven \ac{SC}}: The integration of generative \ac{AI} for synthetic data generation, enriching the \ac{KB} and leveraging \ac{LLM} capabilities for enhanced semantic tasks.

\item 
\textbf{Noise-Augmented Dataset}: A labeled dataset with injected noise, specifically optimized for semantic tasks such as target recognition, localization, and recovery.

\item 
\textbf{Case Study on Semantic Tasks}: A detailed case study that evaluates baseline models across various semantic tasks, assessing performance and adaptability under different noise conditions to validate the GenSC-6G framework.

\end{itemize}

\section{GenSC-6G Dataset Structures}

The GenSC-6G dataset is meticulously organized to support \ac{ML} tasks, including classification, segmentation, object detection, and edge \ac{LLM} tasks. 
Each \ac{ML} or semantic task is associated with a standalone ground-truth data collection or a combination of multiple collections in a generic format. These collections achieve dual objectives: i) they ensure that models can be effectively trained and evaluated across different and interconnected tasks; and ii) the generic format of the dataset provides scalability and flexibility for \ac{SC} tasks, enabling easy modification of wireless methods, parameters, or noise levels. This adaptability allows it to serve goal-oriented tasks in any environment. 
The dataset is available at \texttt{https://github.com/CQILAB-Official/GenSC-6G}.
Each collection of semantic tasks is structured as follows.

\subsubsection{Ground-Truth Data}
The ground truth data collection is precisely annotated for each of the $15$ defined classes, with corresponding class and segmented labels. This dataset includes a total of $4,829$ instances for training and $1,320$ instances for testing. The context of the dataset pertains to common vehicle types in both military and civilian sectors. The input data collection is designed for the semantic upsampling task.

\subsubsection{Base-Model Features}
The dataset includes features extracted from base models in the form of matrices, as illustrated in Fig.~\ref{fig:1}. These extracted features serve as data representations, enabling the processing across multiple decoding components within the large \ac{AI} framework. With these resources, we can enhance the robustness and efficiency of models for goal-oriented tasks.

\subsubsection{Additive Noise Features}
In this component, the dataset includes features with various levels of \ac{AWGN}. These noise features allow for testing the resilience and adaptability of the base models while also enabling in-depth analysis of noise effects and modulation.


\subsubsection{Framework Source Code}
\begin{figure*}[t]
\centering
\includegraphics[width=0.98\linewidth]{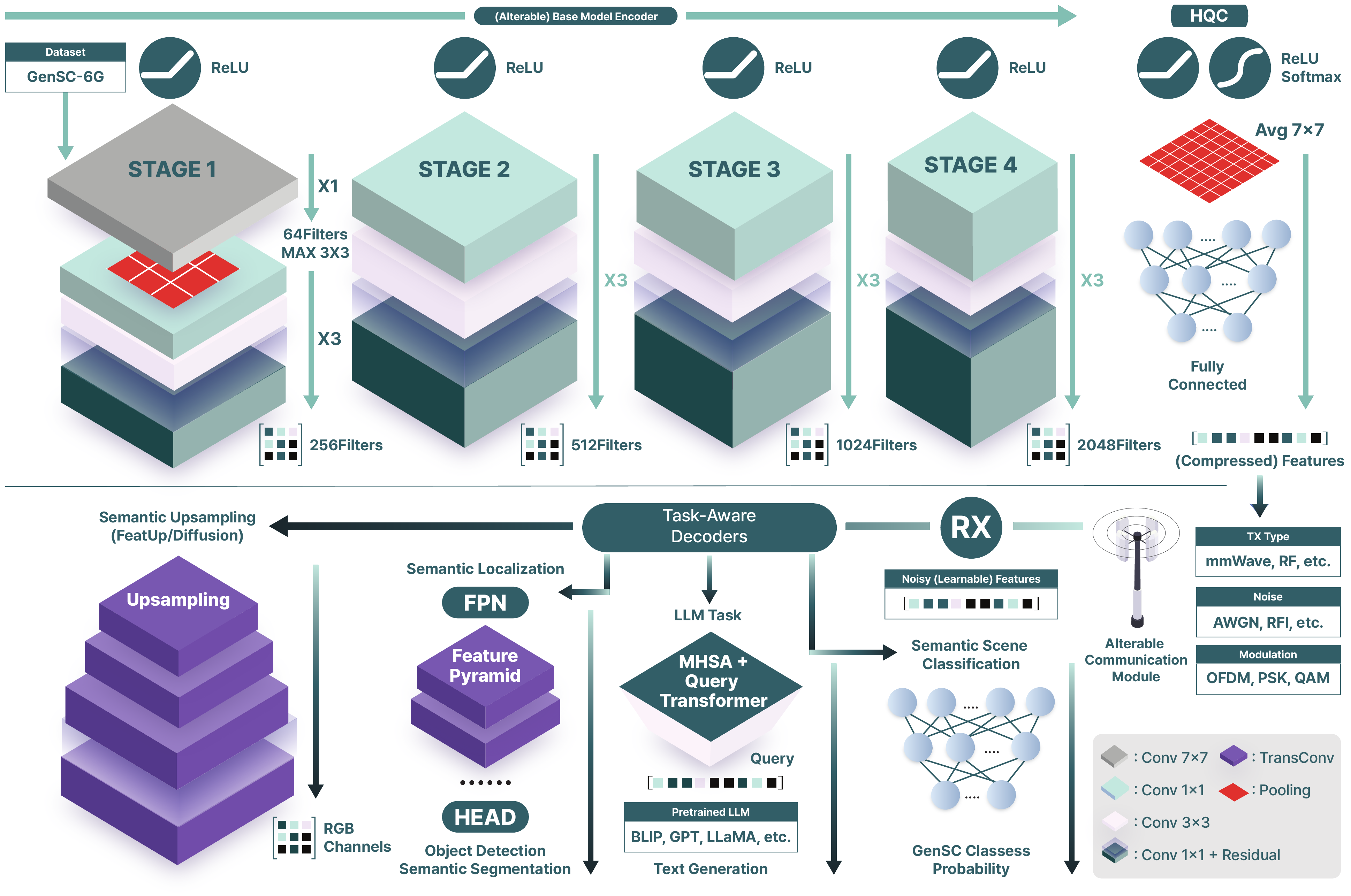}
\caption{
A GenSC-6G testbed prototype.
The large-\ac{AI} \ac{SC} testbed framework prototypes a flexible architecture in which the backbone encoder and communication modules are alterable to fit any backend, and the semantic decoders can be adapted for various downstream goal-oriented tasks.
ReLU, TX, RX, mmWave, OFDM, PSK, QAM, MHSA, BLIP, GPT, LLaMA, and FPN stand for the rectified linear unit, transmitter, receiver, millimeter wave, orthogonal frequency division multiplexing, phase-shift keying, quadrature amplitude modulation, multi-head self-attention, bootstrapping language-image pretraining, generative pretrained transformer, LLM Meta AI, and feature pyramid network, respectively.
}
\label{fig:2}
\end{figure*}

A codebase is provided in the GenSC-6G dataset, including the transmission configuration, base model, and metrics, as illustrated in Fig.~\ref{fig:2}. 
This codebase offers tools for deploying and modifying \ac{SC} models and goal-oriented tasks for customization across various setups.

\section{Prototyping A Large-\ac{AI} SC Testbed}

In this section, we present a modular framework integrating large \ac{AI} models and \ac{HQC} computing with \ac{SC} to enhance data generation, transmission efficiency, and adaptability within the \ac{JSCC} framework. 

\subsection{Generative \ac{AI} Auto Dataset}

We first introduce the concept of dataset auto-creation by integrating diffusion models and automated mechanisms to streamline data generation, labeling, and training processes.

\subsubsection{Diffusion-Driven Data Generator} 

The GenSC-6G testbed leverages text-to-image diffusion models to automate the generation of diverse and trainable images for the base model. This framework uses the \ac{LDM}, which transforms high-dimensional data into low-dimensional latent spaces, optimizing computational efficiency, output quality, and diversity. The \ac{LDM} operates in two main stages: forward diffusion, n which an image is progressively transformed into a noisy representation by adding Gaussian noise over several steps, and reverse diffusion, where this noisy latent representation is iteratively denoised back into a coherent image guided by text prompts. 

\subsubsection{Auto Inference Mechanism} 
With the GenSC-6G dataset pipeline, an auto-generation mechanism is employed in real time to continuously generate new data. The workflow involves feeding the trained model dynamically with input data through a diffusion model, producing diverse data instances. These instances are then labeled automatically and passed to the training pipeline, with the model saved for subsequent training sections. A masked autoencoder model with a \ac{ViT} such as the \ac{SAM} or \ac{YOLO} can be potentially employed for auto-localization tasks (instance segmentation and object detection). This approach allows for consecutive model training without the need for manual data collection.

%
%
%
%

\subsection{Bandwidth-Efficient \ac{SC}}
Efficient compression of semantic data is critical to transmitting relevant information without exhausting resources. 
The model compression mechanism, such as entropy coding, represents the most frequent semantic components with fewer bits, maximizing compression efficiency.
After passing through stages of feature extraction and quantization, where the complexity and number of filters increase at each stage, the data is compressed. This compression is achieved through techniques such as pooling and fully connected layers, which map the extracted features into a compact representation. The compressed data is then encoded into a format suitable for transmission, reducing the burden on available bandwidth by compressing them into latent features.

\subsection{Base-System Model}

We now outline the framework components responsible for encoding, transmission, and task-aware decoding. 

\subsubsection{Feature Extraction} 
In the GenSC-6G framework, the base-system model relies on an alterable backbone encoder that can be replaced with any feature extraction network. This flexibility allows for the use of various \ac{DL} models, such as \acp{CNN} or \acp{ViT}, depending on the target application. Both models extract semantic features from input data but differ in spatial processing: \acp{CNN} use convolutional filters, while \acp{ViT} utilize self-attention mechanisms to capture global context. 
As shown in Fig.~\ref{fig:2}, the backbone encoder begins by passing the input data through several convolutional layers, progressively reducing data dimensionality while retaining key semantic components throughout the patching and filters.

\subsubsection{Semantic Compression} 
The semantic compression process is performed by extracting and prioritizing only the most relevant features from the input data. A critical step of this compression process is quantization. After feature extraction, the data is quantized to map the continuous feature space into a discrete set of values, reducing data precision in a controlled manner. 
The backbone maintains a balance between compression efficiency and semantic relevance preservation. To minimize information loss, a loss function is typically designed to penalize discrepancies between the original data and output, ensuring that the most critical features for downstream tasks are preserved during training.
    
\subsubsection{Pretrained Model} 
The backbone encoders can be conditionally replaced between pretrained and non-trained networks. To accelerate learning, pretrained networks, such as the \ac{ResNet}, \ac{SwinT}, or \ac{ViT}, have been trained on large-scale datasets like ImageNet. By initializing the backbone encoder with a pretrained model, we leverage learned feature representations, improving the convergence speed and sustainability.
    
\subsubsection{Fine-Tuning and Retraining} 
Once the encoder is initialized with a pretrained network, the model can be fine-tuned or retrained. Fine-tuning involves adjusting the pretrained weights with a lower learning rate to adapt to the new task while preserving useful representations---freezing some layers. Retraining involves filling in missing weights, such as those in the decoder and fully connected layers, or resetting the weights entirely---allowing the model to be trained from scratch.
    
\subsubsection{Output} 
After the feature extraction step, the encoder produces a compressed representation of the input data. This compact form reduces data dimensionality while retaining the most significant semantic information, which is then prepared for transmission. The output from the bottleneck can be reused for various goal-oriented tasks, such as classification, object detection, or segmentation.

\subsection{Mobile \ac{DL}}

To accommodate the limited computational resources available on edge devices, the backbone framework is alterable in the optional way to use lightweight neural networks such as EfficientNet-B1 and MobileNet. These models feature  significantly fewer parameters compared to traditional \ac{DL} models, effectively reducing the number of parameters and computational complexity at each layer. As shown in Table~\ref{table:1}, 
this reduction translates into faster inference and load times, which are crucial for real-time applications. EfficientNet-B1 and MobileNet achieve this by optimizing network layers, employing depthwise separable convolutions, and reducing the overall model size, all while striving to maintain high accuracy with lower computational demands.

\subsection{Quantum Parallel Processing}

\Ac{HQC} computation in the large-\ac{AI} \ac{SC} testbed framework combines \acp{QPU} with traditional \acp{GPU} to execute \ac{DL} tasks more efficiently. \acp{QPU} work alongside \acp{GPU} to handle tasks such as \ac{HQC} optimization, state preparation, and sampling, which complement \ac{GPU} operations on matrix multiplication and dense processing. The testbed framework employs skip connections and resource splitting, enabling \acp{QPU} to manage quantum-specific computations while \acp{GPU} process standard neural network layers. Specifically, to implement this, the model architecture is enhanced with quantum layers, consisting of a basic entanglement layer and amplitude embedding. The skip connection features are conditionally encoded into these quantum layers for \acp{QPU}. In general, this task division helps manage computational loads on backbone feature extraction, decoding, and inference.
Hence, the framework incorporates quantum kernels, which introduce quantum properties into the learning process to enhance model performance. These quantum kernels, in the form of embeddings within \acp{QPU}, map data into high-dimensional quantum Hilbert spaces, providing more powerful data representations that can improve classification accuracy and decision-making processes.  This hybrid approach allows the system to dynamically offload complex tasks to cloud-based \acp{QPU} when on-premises hardware reaches its computational limits, ensuring scalability with future advancements in processing unit technology.


\begin{figure}[t]
\centering
\includegraphics[width=0.98\linewidth]{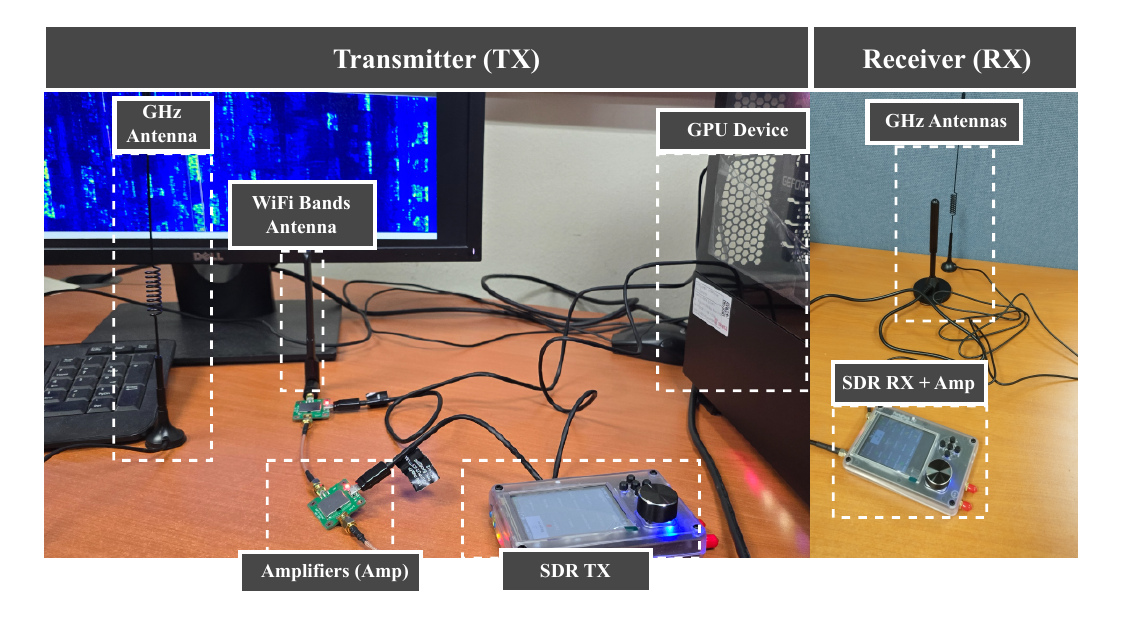}
\caption{
A transceiver setup to capture noise features as part of the testbed. The testbed leverages the Wi-Fi~$7$ ($802.11$be) OFDM communication system with file streaming. On the transmitter side, a programmable SDR setup with gigahertz (GHz) antennas and amplifiers sends high-frequency signals to the receiver. 
}
\label{fig:3}
\vspace{-10pt}
\end{figure}

\begin{table*}
    \caption{
    Performance of large-\ac{AI} base models trained on the GenSC-6G dataset for classification, upsampling, and edge \ac{LLM} tasks.
    }
    
    \begin{center}
    
    \begin{tabular}{l l l l l c c c c c c} 
    
    \toprule
    
    \multirow{3}{*}{Device} 
    & 
    \multirow{3}{*}{\makecell[l]{Backbone\\Encoder}} & 
    Decoder &
    \multirow{3}{*}{\makecell[l]{Parameters\\(Upsamplers)}} & 
    \multirow{3}{*}{\makecell[l]{Processing\\Unit}} & 
    \multicolumn{3}{c}{AWGN ($\text{SNR}=10$\,dB)} & 
    \multicolumn{3}{c}{AWGN ($\text{SNR}=30$\,dB)}  \\
    \cmidrule(lr){3-3}  \cmidrule(lr){6-8} \cmidrule(lr){9-11}
    & & Classifier & & & Accuracy & F1 & Recall & Accuracy & F1 & Recall  \\
    & & (Upsampler) & & & (LPIPS) & (CLIP-S$^\ast$) & (CLIP-S$^\dag$) & (LPIPS) & (CLIP-S$^\ast$) & (CLIP-S$^\dag$)  \\

    \midrule
    \midrule
    
    \multirow{7}{*}{Classical} 
    &    
    ViT-L-32 & 3xFC & $306.79$ &GPU & 0.8477 & 0.8514 & 0.8512 & 0.8485 & 0.8522 & 0.8516   \\
    
    &
    ViT-L-32  & (FeatUp) &  $(30.75)$ &GPU & (0.4211) & (28.0641) & (29.9747) & (0.4038) & (27.5533) & (30.0994)   \\
    
    &
    ResNet-50 & 3xFC & $25.81$ & GPU
     & 0.8447 & 0.8468 & 0.8462 & 0.8485 & 0.8507 & 0.8510   \\
    
    &
    ResNet-50 & (FeatUp) &  $(30.75)$ & GPU  & (0.4163) & (27.7008) & (29.9923) & (0.4027) & (27.9054) & (30.0679)   \\
    
    &
    VGG-16 & 3xFC & 138.61 & GPU & 0.8144 & 0.8163 & 0.8158 & 0.8167 & 0.8158 & 0.8161   \\
    
    &
    Inception-V3 & 3xFC & 27.42 & GPU & 0.8561 & 0.8569 & 0.8553 & 0.8644 & 0.8650 & 0.8641 \\
    
    &
    DINO-V2  & (FeatUp) &  $(30.75)$ &GPU & (0.4210) & (27.6070) & (29.8617) & (0.4153) & (27.7631) & (30.0411)   \\

    \midrule
    
    \multirow{2}{*}{Mobile} 
    
    &
    EfficientNet-B1 & 2xFC & 7.79 & CPU & 0.8689 & 0.8702 & 0.8700
     & 0.8705 & 0.8720 & 0.8735\\
    
    &
    MobileNet-V3 & 2xFC & 5.48 & CPU & 0.7871 & 0.7889 & 0.7896 & 0.8197 & 0.8365 & 0.8192   \\
    
    \midrule
    
    \multirow{2}{*}{HQC} 
    
    &
    ViT-L-32 & QNN & 306.79 & GPU/QPU & 0.8303 & 0.8250 & 0.8232 & 0.8485 & 0.8522 & 0.8516   \\
    
    &
    ResNet-50 & QNN & 25.81 & GPU/QPU & 0.8144 & 0.8181 & 0.8185 & 0.8356 & 0.8383 & 0.8381  \\
     
    
    %
    %
    %
    %

    \bottomrule
    
     \multicolumn{7}{l}{Note: $^\ast$\,LLaMA-3; $^\dag$\,BLIP-2}   &  &  &    \\

    \end{tabular}
    \label{table:1}
    \vspace{-20pt}
    \end{center}
    \end{table*}

\subsection{Classical and Quantum \ac{JSCC} Modules}
The GenSC-6G \ac{JSCC} module is designed to evaluate the quality degradation and performance of data transmission under both classical and quantum channel conditions.

\subsubsection{Classical Channel}
The classical channel in the GenSC-6G prototype is configured to emulate realistic communication scenarios, particularly focusing on noise conditions, modulation schemes, and transmission protocols representative of current and next-generation communication standards.

\begin{itemize}
    
\item 
\textbf{Channel Noise:} 
Noise is implemented within the testbed to replicate real-world communication scenarios where data transmission is disturbed. In this case, the system model simulates noise conditions by incorporating \ac{AWGN} with a specific \ac{SNR} and random transmitter noise from a programmable \ac{SDR} to dynamically adjust noise characteristics. An example setup is shown in Fig.~\ref{fig:2}. By injecting controlled and random noise levels during transmission, the system can assess the impact on semantic data, helping to refine error correction and noise mitigation within the network decoder. The \ac{DL} model adapts to varying noise conditions by optimizing its parameters to minimize the loss function, which measures the discrepancy between the predicted noisy outputs and the ground-truth data.

\item 
\textbf{\ac{JSCC} Transmission:} 
The prototype utilizes \ac{JSCC} to efficiently transmit semantic information over bandwidth-limited channels. 
By combining source compression and channel coding within a single \ac{DL} framework, the system learns an end-to-end mapping from input data to transmitted signals and from received signals to reconstructed outputs. In this process, the semantic features are directly mapped into channel symbols, bypassing traditional separate source and channel coding schemes.
The testbed transmission employs \ac{OFDM} with Wi-Fi~$7$ ($802.11$be) (see Fig.~\ref{fig:3}). 
The system can be adapted to various frequency bands, including \ac{THz} and sub-\ac{THz} bands, which provide extensive bandwidth for high-speed semantic applications. 
    
\end{itemize}

\subsubsection{Quantum Channel}
In the quantum communication scenario, the \ac{JSCC} module is adapted to encode semantic information into quantum states for transmission over quantum channels that exploit quantum properties. The typical integration in quantum transmission can be described as follows.

\begin{itemize}

\item 
\textbf{Classical-Quantum Embeddings:} 
This process maps the semantic features extracted from the backbone encoder into quantum states. 
The features are first passed through a fully connected layer for dimensionality reduction, preparing them for quantum encoding. The prototype can then utilize amplitude-embedding and angle-embedding techniques to encode these features into qubits.
In amplitude embedding, the normalized feature vector is directly used to define the amplitudes of the quantum state, allowing for efficient representation of high-dimensional data in a quantum system. This method maps the classical information into the amplitudes of a superposed quantum state. In angle embedding, the features are encoded into the rotation angles of quantum gates, such as parameterized rotation gates on the Bloch sphere, which manipulate the state of the qubits accordingly. These embeddings translate classical data into quantum formats suitable for quantum processing and transmission. Additionally, an entanglement layer can be incorporated into the quantum encoder to exploit the entanglement properties. This layer uses quantum gates, such as the controlled-NOT or controlled-phase gates, to establish entanglement among qubits, thereby enabling the quantum system to capture complex feature interactions and dependencies within the semantic data.

\item 
\textbf{Bit-Flip and Phase-Flip Noise:} 
In quantum communication, qubits are susceptible to various types of noise, including bit-flip and phase-flip errors. Bit-flip noise alters the state of a qubit from $\ket{0}$ to $\ket{1}$ or vice versa, while phase-flip noise changes the phase between states without affecting their probabilities. These noise types are typically modeled using Pauli operators, specifically the Pauli-X (bit-flip) and Pauli-Z (phase-flip) gates. Quantum error correction codes, such as Shor or Steane codes, mitigate these errors by leveraging entanglement and redundancy.
\end{itemize}



\begin{figure*}[t]
    \centering
\includegraphics[width=0.98\linewidth]{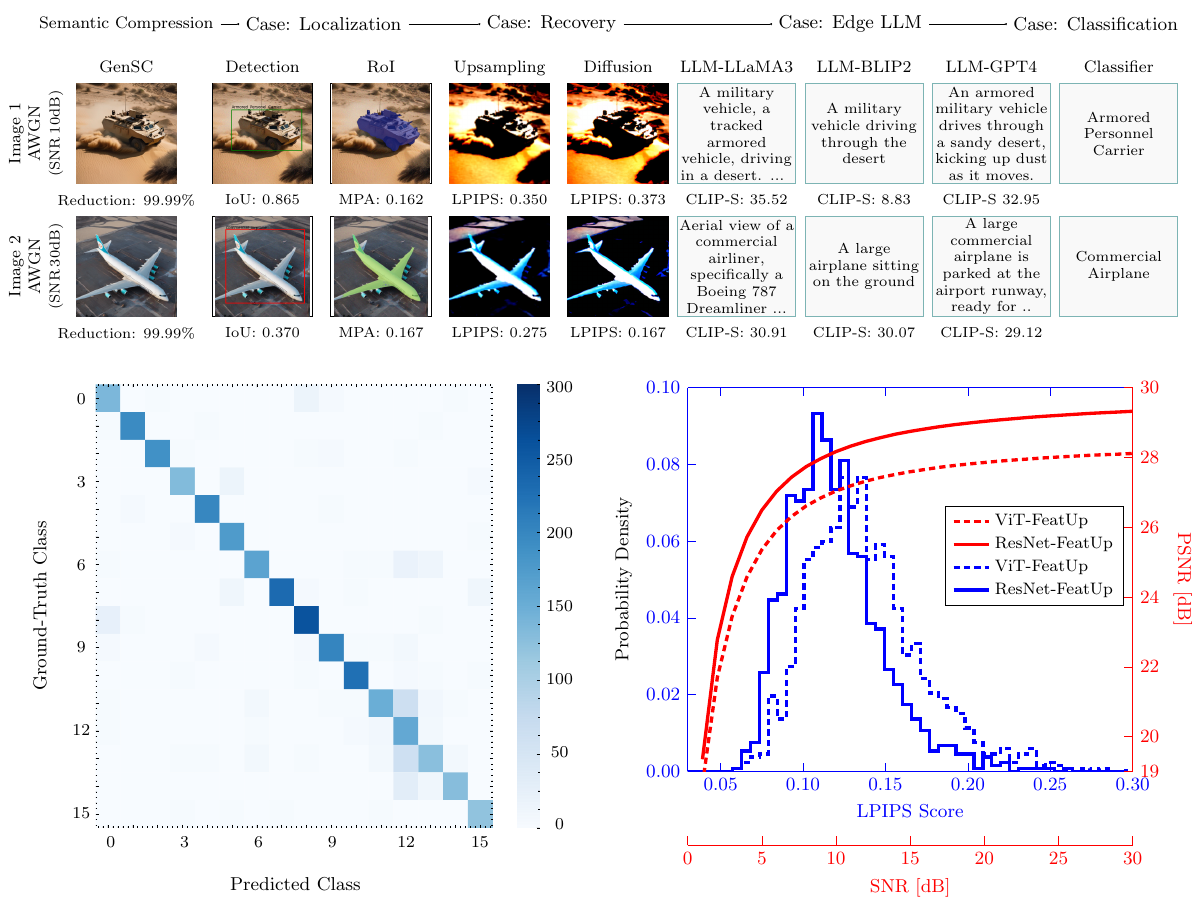}
    \caption{Overview of case studies demonstrating downstream goal-oriented tasks from encoded features and their common metrics, including semantic compression, object localization, recovery through upsampling and diffusion, and post-processing with \ac{LLM}, illustrating the adaptability of encoded features. Herein, the average compression rate of the dataset reduces to $99.993\%$, which is compressed to nearly zero. The confusion matrix (bottom left) illustrates the classification performance of the ResNet-50 model under different SNR conditions. The LPIPS probability distribution and the PSNR (bottom right) contrast perceptual similarity scores between different models and reflect image quality maintenance across varying noise levels, showing the adaptability and effectiveness of these models.
    }
    \label{fig:4}
\end{figure*}


\subsection{Case Study} 

We now present case studies demonstrating the application of the GenSC-6G prototype to various semantic decoding tasks. Each task utilizes the compressed semantic features transmitted over the communication channel and focuses on different aspects of semantic understanding. We evaluate the performance using relevant metrics and provide benchmarks for comparison.

\subsubsection{Lightweight Classification}

The semantic classification task involves categorizing images into predefined classes based on their content. Using the GenSC-6G dataset, we train several baseline models, including \ac{ViT}-L-32, \ac{ResNet}-50, \ac{VGG}-16, and Inception-V3, on single or combined processing units, as well as lightweight models like EfficientNet-B1 and MobileNet-V3 suited for edge devices. All baseline models are trained under various SNR conditions (e.g., at $10$\,dB and $30$\,dB) to evaluate their robustness to noise in the communication channel. The decoder architecture is defined as a lightweight module with three fully connected layers designed to downsample the features extracted by the encoder.
Fig.~\ref{fig:4} shows the confusion matrix for the \ac{ResNet}-50 model under \ac{AWGN} with the \ac{SNR} of $10$\,dB (lower-left plot), illustrating the balance between true and false positive rates. The model achieves an accuracy of $84.47\%$ (see Table~\ref{table:1}), demonstrating robust---but improvable---baseline performance in noisy conditions. Notably, the highest baseline accuracy is achieved by EfficientNet-B1 with a score of $86.89\%$ while maintaining a mobile-friendly architecture. Additional metrics, including F1 score and recall, are also provided in Table~\ref{table:1}. By focusing on transmitting essential semantic features and utilizing lightweight decoders, the models demonstrate that high classification accuracy is achievable even in the presence of significant channel noise. 
Furthermore, these compressed features are not only effective for classification tasks but are also reusable for other semantic decoding tasks in this case study, as they retain rich semantic representations of the input data. This reusability highlights their efficiency for multiple downstream \ac{AI} tasks.

\subsubsection{Semantic Localization}
We first utilize the \ac{YOLO} model for object detection of vehicles within the images from the GenSC-6G dataset. After detecting the vehicles with \ac{YOLO}, we perform semantic segmentation to achieve pixel-level localization. The ground truth provided by the GenSC-6G segments is used to train and validate the segmentation models. By leveraging these detailed annotations, the models learned to accurately delineate vehicle boundaries, enhancing the localization precision.
As shown in Fig.~\ref{fig:4} (upper plot), we use localization metrics such as the \ac{IoU} and \ac{MPA}. The higher \ac{IoU} for Image~$1$ at the \ac{SNR} of $10$\,dB indicates better overlap between the predicted segmentation and the ground truth, suggesting accurate vehicle localization by the model in that instance.
The \ac{MPA} values, while relatively low, provide insight into pixel-level classification accuracy across the entire image, including both the object and the background. The close \ac{MPA} values between the two images indicate consistent pixel-level performance, though there is room for improvement, especially in challenging conditions. These results demonstrate that the \ac{SC} framework effectively supports semantic localization tasks by preserving essential spatial features necessary for accurate object detection and segmentation. 

\subsubsection{Semantic Upsampling Recovery}
The semantic upsampling recovery task focuses on enhancing low-resolution images received over noisy channels by reconstructing high-resolution outputs. Two distinct approaches are evaluated for upsampling combined with feature upsampling (FeatUp) for two baseline models (\ac{ResNet} and \ac{ViT}). FeatUp enhances deep features by restoring lost spatial information through high-resolution signal guidance or implicit modeling to improve performance in dense prediction tasks.
To ascertain upsampling recovery performance, we evaluate the \ac{LPIPS} and the peak \ac{SNR} (\acs{PSNR}) for the GenSC-6G dataset in Fig.~\ref{fig:4} (lower-right plot). As depicted, there is an inverse relationship between the probability of accurate reconstruction and the \ac{LPIPS} score. Lower \ac{LPIPS} values indicate that reconstructed images are perceptually closer to the ground truth. The empirical distribution demonstrates that the \ac{ResNet} with FeatUp has a higher probability density in the lower LPIPS score range (from $0.05$ to $0.15$), indicating more accurate reconstructions, while the \ac{ViT} baseline with FeatUp remains notable performance.
The \ac{PSNR} performance is also depicted as a function of \ac{SNR}. Here, \ac{ResNet}-FeatUp outperforms \ac{ViT}-FeatUp again, especially at high \ac{SNR} values, demonstrating considerable noise resilience and image fidelity during upsampling recovery. 

\subsubsection{Edge \ac{LLM}}
The edge \ac{LLM} task involves integrating advanced models, such as the \ac{BLIP}, \ac{GPT}, and \ac{LLaMA}, into edge environments to generate text from semantic features extracted from visual inputs, enriching the output semantics. As depicted in Fig. \ref{fig:2}, the prototype architecture employs a feature encoder combined with a querying transformer. These queries are then input into pretrained \acp{LLM} that specialize in transforming visual-semantic representations into meaningful text outputs. The visual-text encoder architecture plays a pivotal role in this setup, combining visual feature extraction with a text generation pipeline. 
The \ac{CLIP-S} is used to measure the alignment between the generated text and the visual context (see Fig.~\ref{fig:4} upper plot). The \ac{CLIP-S} evaluates how well the descriptions or captions match the visual input, reflecting the effectiveness of generated text in comprehending the scene. For example, in GenSC-6G image captioning, \ac{LLaMA}-3 reaches a \ac{CLIP-S} of $35.52$ for Image~$1$, demonstrating its level of contextual understanding and the richness of the text generated from the image.

\balance

\section{Open Challenges and Conclusion}
Despite the results demonstrated by the GenSC-6G framework, several challenges remain in fully realizing its potential. Sustainability is a significant concern, particularly in maintaining energy-efficient operations for large models and continuous data processing at the edges. This is critical for ensuring that \ac{AI}-driven \ac{SC} systems align with the sustainability goals of future networks. Model robustness in the face of varying noise conditions and unpredictable environments also remains challenging.
Deploying these models on mobile devices poses another challenge, as many of the state-of-the-art \ac{AI} models, including \acs{LLM}, are computationally intensive and difficult to scale down to the limited resources of edge devices. Solutions such as model pruning, quantization, and efficient architecture designs require further refinement to enable real-time on-device processing. Generalization to more downstream \ac{AI} tasks, including complex multimodal tasks, is also crucial to expand the SC utility. 
Finally, quantum communication and computing face challenges in ensuring noise resilience, enhancing scalability, and managing the complexity of parallel processing.

This paper has introduced the GenSC-6G framework, which integrates large \ac{AI}, \ac{HQC} optimization, and \ac{SC}, tailored to optimize 6G networks through scalable and alterable communications models. The testbed offers a flexible prototype that enables the modification of baseline models, communication modules, and goal-oriented decoders, supporting a variety of downstream tasks. This modular framework leverages generative \ac{AI} to enhance the \ac{KB} by generating realistic synthetic data, thus improving model diversity and adaptability in real-world scenarios.
Through the detailed case studies, we have demonstrated the effectiveness of our approach in various semantic decoding tasks, including lightweight classification, semantic localization, and upsampling recovery. Evaluations across different communication conditions, as seen in downstream tasks such as semantic classification and edge \ac{LLM}, highlight the practical adaptability of the GenSC-6G framework in a wide range of semantic tasks.


\bibliographystyle{IEEEtran}
\bibliography{
IEEEabrv, 
CQILAB-abrv,
CQILAB-Journal,
CM-GENSC
}

\vspace{0.45cm}
\footnotesize
\noindent
\textbf{Brian E. Arfeto} is a Ph.D. student at Kyung Hee University, Korea.\\

\vspace{-0.05cm}
\noindent
\textbf{Shehbaz Tariq} is a Ph.D. student at Kyung Hee University, Korea.\\

\vspace{-0.05cm}
\noindent
\textbf{Uman Khalid} is a postdoctoral fellow at Kyung Hee University, Korea.\\

\vspace{-0.05cm}
\noindent
\textbf{Hyundong Shin} [F] is a professor at Kyung Hee University, Korea.\\

\vspace{-0.05cm}
\noindent
\textbf{Trung Q. Duong} [F] is a professor at Memorial University, Canada and Queen's University Belfast, U.K.\\

\end{document}